\documentclass[conference]{IEEEtran}
\IEEEoverridecommandlockouts
\usepackage{cite}
\usepackage{amsmath,amssymb,amsfonts}
\usepackage{algorithmic}
\usepackage{graphicx} 
\usepackage{times}
\usepackage{latexsym}
\usepackage{booktabs} 
\usepackage{tikz} 
\usepackage{array}
\usepackage{textcomp}
\usepackage{hyperref}
\usepackage{xcolor}
\def\BibTeX{{\rm B\kern-.05em{\sc i\kern-.025em b}\kern-.08em
    T\kern-.1667em\lower.7ex\hbox{E}\kern-.125emX}}
\begin{document}

\title{Multi-Agent LLMs for Generating Research Limitations}

\author{\IEEEauthorblockN{Ibrahim Al Azher}
\IEEEauthorblockA{\textit{Department of Computer Science} \\
\textit{Northern Illinois University}\\
DeKalb, IL, USA \\
iazher1@niu.edu}
\and
\IEEEauthorblockN{Zhishuai Guo}
\IEEEauthorblockA{\textit{Department of Computer Science} \\
\textit{Northern Illinois University}\\
DeKalb, IL, USA \\
zguo@niu.edu}
\and
\IEEEauthorblockN{Hamed Alhoori}
\IEEEauthorblockA{\textit{Department of Computer Science} \\
\textit{Northern Illinois University}\\
DeKalb, IL, USA \\
alhoori@niu.edu}}

\maketitle

\begin{abstract}
Identifying and articulating limitations is essential for transparent and rigorous scientific research. However, zero-shot large language models (LLMs) approach often produce superficial or general limitation statements (e.g., dataset bias or generalizability). They usually repeat limitations reported by authors without looking at deeper methodological issues and contextual gaps. This problem is made worse because many authors disclose only partial or trivial limitations. We propose, a multi-agent LLM framework for generating substantive limitations. It integrates OpenReview comments and author-stated limitations to provide stronger ground truth. It also uses cited and citing papers to capture broader contextual weaknesses. In this setup, different agents have specific roles as sequential role: some extract explicit limitations, others analyze methodological gaps, some simulate the viewpoint of a peer reviewer, and a citation agent places the work within the larger body of literature. A Judge agent refines their outputs, and a Master agent consolidates them into a clear set. This structure allows for systematic identification of explicit, implicit, peer review-focused, and literature-informed limitations. Moreover, traditional NLP metrics like BLEU, ROUGE, and cosine similarity rely heavily on n-gram or embedding overlap. They often overlook semantically similar limitations. To address this, we introduce a pointwise evaluation protocol that uses an LLM-as-a-Judge to measure coverage more accurately. Experiments show that our proposed model substantially improve performance. The RAG + multi-agent GPT-4o mini configuration achieves a +15.51\% coverage gain over zero-shot baselines, while the Llama 3 8B multi-agent setup yields a +4.41\% improvement. Code and datasets are available at: code
\url{https://github.com/IbrahimAlAzhar/LimAgents_limitation_generation_with_LLM_Agents} | Dataset: \url{https://huggingface.co/datasets/iaadlab/LimAgents_limitation_data_scientific_papers_with_cited_papers} 

\end{abstract} 

\begin{IEEEkeywords}
Research limitations, Text Generation, LLM Agents, multi-agent systems, retrieval-augmented generation (RAG), Peer review, Evaluation methods
\end{IEEEkeywords}

\section{Introduction} 
Limitations of a scientific article refer to restrictions in study design, methodological flaws, or shortcomings that affect the interpretation, validity, or generalizability of the findings \cite{b1}. Scholars emphasize that acknowledging limitations is essential for transparent reporting, as they clarify the boundaries within which results should be understood. Identifying limitations in scientific manuscripts is thus a crucial step in both manuscript preparation and critical appraisal. Despite their importance, limitations are often underreported or downplayed. Ioannidis \cite{b18} famously showed that only 17\% of top-tier articles mention limitations in the main text and a mere 1\% in abstracts or dedicated sections. In biomedical research, 27\% of papers omit any discussion of limitations altogether \cite{b19}. In addition, authors sometimes provide generic or irrelevant limitations that fail to convey the study’s real constraints \cite{b7}, or use them as a form of hedging, presenting minor weaknesses while avoiding deeper flaws to reduce the risk of rejection \cite{b29}. Contributing factors include limited awareness of the importance of limitations and, more critically, fear that openly acknowledging major weaknesses could lead to rejection from journals or conferences, since reviewers may perceive a large number of significant limitations as undermining the paper’s contribution \cite{b7}. This often results in limitations sections that are superficial, incomplete, or strategically downplayed. 
By highlighting weaknesses that are often overlooked in manuscripts, limitation generation improves the transparency of scientific reporting and reinforces the critical evaluations central to peer review. Moreover, the rapid growth of submissions has further strained the peer review process. For example, ICLR submissions increased by 47\% in 2024 and 61\% in 2025, which has been linked to reviews that are frequently brief, vague, or unprofessional \cite{b40}. In this context, there is a need for tools that act as aids to the review process, helping prevent overlooked weaknesses, promoting consistency, and allowing reviewers and editors to focus on substantive issues rather than surface-level remarks. Moreover, some top-tier conferences have begun experimenting with LLMs in the review workflow (e.g., ICLR’25 \cite{b3}, AAAI’25 \cite{b4}). 
However, despite their promise, LLM-based peer review systems exhibit several shortcomings. For example, evaluations of ICLR 2024 submissions found that LLM reviewers hallucinated details, favored verbose papers, and exhibited institution-based biases \cite{b24}. In some cases, even empty submissions received positive and detailed feedback, while longer papers were consistently rated more favorably. Single-blind settings also led to biases in favor of prestigious institutions \cite{b23}. 

Generating limitations plays a crucial role in the peer review process, as reviewers are expected to critically assess both the strengths and weaknesses of a study. But what is the problem with LLM-generated limitations in peer review? When prompted in a zero-shot setting, LLMs often fall back on vague and generic critiques, such as dataset bias or generalization issues, or simply reproduce author-stated limitations \cite{b8}. These outputs overlook deeper, implicit weaknesses and lack the nuance required for rigorous scientific assessment \cite{b6}. To address these shortcomings, we propose an effective LLM agent-based framework to generate a more complete and meaningful set of limitations for scholarly articles—supporting their critical appraisal within digital libraries—without requiring specialized training or significant computational expense. Instead of relying on a single-pass zero-shot prompt, our approach decomposes the task into specialized agents with distinct roles. In this paper, we use the term \textit{LLM agents} to refer to LLM instances configured with specific responsibilities that collaborate within a multi-agent workflow. The \textbf{Extractor Agent} identifies verbatim author-stated limitations. The \textbf{Analyzer Agent} performs a methodological audit to infer implicit shortcomings such as sample bias or statistical issues. The \textbf{Reviewer Agent} adopts a peer-review stance to highlight concerns about reproducibility, transparency, or ethics. The \textbf{Citation Agent} incorporates contextual limitations from cited and citing works, capturing broader methodological or baseline weaknesses that may not be explicitly acknowledged in the target paper. To automate our data extraction pipeline, our LLM agents incorporated external tools to collect papers and extract their content. These agents are complemented by oversight and refinement mechanisms. Moreover, a \textbf{Judge Agent} evaluates outputs across multiple dimensions (faithfulness, soundness, importance, depth, originality, actionability, and topic coverage), providing targeted feedback when necessary. If an agent’s performance falls below a threshold, a \textbf{Self-Feedback Agent} prompts regeneration to improve quality. Finally, a \textbf{Master Agent} consolidates all outputs, resolves redundancies, and produces a cohesive, de-duplicated list of limitations labeled with their provenance (“Author-stated,” “Inferred,” “Peer-review–derived,” or “Cited”).  

In addition, to evaluate performance, we move beyond traditional metrics such as BLEU, ROUGE, or cosine similarity, which primarily measure n-gram or embedding overlap and underestimate coverage when phrasing differs. Instead, we introduce a \textbf{pointwise evaluation} approach, where each generated limitation is compared against a reference at the topic level. Moreover, to construct a stronger reference set, we combine author-stated limitations with peer-review comments from OpenReview, since relying only on author disclosures risks overlooking substantive weaknesses. Taken together, our framework combines multi-agent decomposition, broader contextual evidence, iterative self-refinement, and robust evaluation to generate limitations that are both more nuanced and more impactful than those produced by zero-shot prompting. Our research questions are threefold, focusing on the effectiveness of LLM agents, their optimal configuration, and the evaluation strategies required to measure their performance:
\begin{itemize} 
\item \textbf{RQ1}: Can LLM agents outperform zero-shot prompting in both smaller and larger models when generating scientific article limitations?  
\item \textbf{RQ2}: What is the optimal agent configuration (number and role of agents) for effectively generating limitations, and how does agent decomposition improve performance compared to a single-pass approach? 
\item \textbf{RQ3}: How reliable are different evaluation strategies including coverage, NLP-based metrics, and LLM-based judgments in capturing the quality and diversity of generated limitations? 
\end{itemize}

\section{Related Work}

Earlier research has analyzed various components of scientific articles, such as abstracts, introductions, datasets, methodologies \cite{b11}, acknowledgments \cite{b10}, and future work \cite{b12}. More recently, greater attention has been given to the automated analysis and generation of limitations. Researchers have explored several methods, including retrieving limitation passages directly from papers \cite{b16}, synthesizing limitations from topic modeling with LLM \cite{b13}, building new benchmark datasets \cite{b17}, and generating limitations for visual elements like charts and graphs \cite{b14}. Recent approaches have also focused on applying zero-shot prompting to LLMs \cite{b39} and developing comprehensive frameworks to benchmark RAG-based generation with novel evaluation methods \cite{b1}.

\textbf{Peer Review.} Recent work has explored zero- and few-shot LLMs for paper rating and end-to-end review generation \cite{b22}. Along these lines, several systems have been proposed to improve review generation. For example, OpenReviewer \cite{b31} fine-tunes a domain-specific LLM (Llama-8B) on large-scale review datasets to produce structured, venue-aligned feedback. DeepReview \cite{b33} introduces a multi-stage reasoning framework to mitigate hallucinations and strengthen evidence-based critiques, while ReviewRobot \cite{b34} leverages knowledge graphs from papers and citations to generate explainable comments. Other studies have benchmarked GPT-based reviewers against humans \cite{b35,b36}, generally finding overlap but also highlighting biases, shallow critiques, and inconsistencies. These works underscore both the promise and the shortcomings of LLM-based peer review, reinforcing the need for systematic approaches to tasks. 


\textbf{LLM Agents.} Building on these efforts, researchers have begun exploring multi-agent approaches to improve the quality and reliability of review assistance. Prior work has demonstrated the utility of role-based or multi-step frameworks \cite{b25}. More recently, multi-agent systems have been applied to peer review \cite{b41}. AgentReview modeled reviewers, authors, area chairs, and meta-reviewers across multiple phases. Ye et al. \cite{b23} employed a pipeline where vague or generic review comments were refined through multiple stages. However, current LLM agents face notable challenges: they can suffer from specification issues, inter-agent misalignment, task verification errors, and poorly structured feedback \cite{b21}. These challenges suggest that although multi-agent setups show potential, building such an effective system requires careful agent design and a mechanism.

Prior work has explored using LLMs and RAG for limitation generation. To the best of our knowledge, our work is the \textbf{first to systematically investigate this gap by applying LLM agents to limitation generation in scientific articles}. We experiment with different agent types and configurations, introduce a pointwise evaluation method, and demonstrate how multi-agent decomposition can generate more meaningful and contextually grounded limitations than zero-shot prompting.

\section{Data Collection} 
\textbf{Data Collection Agent.}  To construct our dataset, we deployed a Data Collection Agent tasked with assembling a corpus for 2,700 NeurIPS papers from 2022 and 2023. This agent first integrates with ScienceParse\footnote{https://github.com/allenai/science-parse} to parse the full text of each primary paper. The agent also utilizes an integrated, Selenium-based component to scrape and link the official OpenReview comments associated with each paper.
Once the raw texts were gathered, our pipeline proceeded with a two-stream extraction process for author-stated and reviewer-stated limitations.
For author-stated limitations, we first applied a rule-based approach to the main paper text. This method scanned for the keyword “limitation” and captured all subsequent text until the next section heading, excluding sections like the abstract, introduction, and related work. The process would also stop if terminal keywords such as ``grant," ``acknowledgments," or ``future work" were encountered. As this initial pass often contained noisy, unrelated sentences, the extracted text was then sent to an LLM for refinement, which re-extracted only the precise limitation-related content.
For reviewer-stated limitations, we processed the parsed OpenReview text. The LLM, configured with zero temperature for consistency, was used to extract all passages referring to weaknesses or critiques from the feedback.
Finally, a dedicated merger step was applied by LLM to combine and de-duplicate the limitations from both streams (the author-stated text and the OpenReview comments), creating a final, unified set for each paper.
On average, each paper's text contains 6–7 limitations, while its OpenReview discussions yield an additional 10–12 limitations. In total, our final dataset comprises 51,300 limitations across the 2,700 research papers.

\textbf{Human Evaluation:} We conducted a human evaluation to verify that the LLM was extracting limitations rather than generating new content. Annotators answered three Yes/No questions: (1) Does the LLM extract limitations from the research paper without generating them itself? (2) Does the LLM extract limitations from OpenReview comments rather than generating them? (3) Does the LLM merge author-stated and OpenReview limitations without introducing new information? We randomly selected 500 samples and recruited three graduate students with ML/NLP expertise. All annotators answered “Yes” to all questions, confirming that the LLM effectively extracts limitations from research papers and OpenReview without hallucinating additional content.

\section{Methodology} 
\begin{figure}[htbp]
    \centering
      \includegraphics[width=\textwidth,height=0.40\textheight,keepaspectratio]{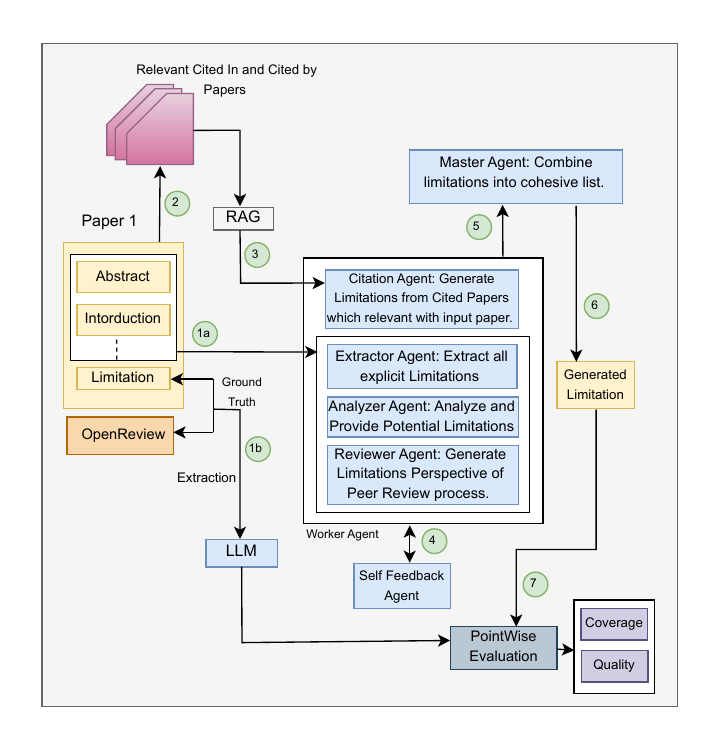} 
     \vspace{-3pt}
    \caption{Workflow of our Proposed Framework.}
    \label{fig:arch}
    \vspace{-10pt}  
\end{figure}

Figure \ref{fig:arch} illustrates our proposed framework for generating and evaluating limitations. For any given paper, our system uses two parallel streams to generate an initial set of limitations. Our model consists with multiple LLM agents with sequential approach. First, Extractor, Analyzer, and Reviewer agents process the paper's main text (Step 1a). In parallel, a Citation Agent uses a RAG system populated with cited and citing papers to generate additional, context-aware limitations (Step 3). A Self-Feedback Agent then evaluates the outputs from all four worker agents (Extractor, Analyzer, Reviewer, and Citation). If any generated limitation falls below a quality threshold, it is sent back to the respective agent for re-generation (Step 4). Finally, a Master Agent aggregates and de-duplicates all validated outputs into a cohesive set (Step 5). This final set is then compared against a ground truth, which we construct from author-stated limitations and OpenReview comments, using a pointwise evaluation method (Steps 1b \& 7) (Prompts Figure \ref{fig:prompt-llm-agents}, Detailed prompts in \href{https://github.com/IbrahimAlAzhar/LimAgents_limitation_generation_with_LLM_Agents/blob/main/Prompt/LimAgents_Prompt.ipynb}{code})

\subsection{Extractor Agent} 

Authors often mention explicit shortcomings outside the designated Limitations section, for example, in the discussion or conclusion sections. Motivated by this, our Extractor Agent is designed to capture these statements wherever they appear. The agent’s input is the full paper text, excluding the ground-truth limitation, thereby tasking it with identifying all other explicit limitations scattered throughout the manuscript. Inspired by the chain-of-thought paradigm \cite{b2}, the agent follows a ``chain of limitations" process. It first scans high-probability sections, searching for explicit cues such as “a limitation of this study.” The agent then extracts the verbatim statements of these self-reported weaknesses (e.g., sample size constraints or methodological caveats). To ensure the output faithfully reflects the authors’ own acknowledgments, the agent strictly avoids inference or external interpretation. Each finding is then formatted into a structured object containing the verbatim text and its source location. Finally, the agent cross-verifies the extraction against the source text for accuracy. By design, this agent focuses only on explicitly declared shortcomings, leaving the task of identifying implicit weaknesses to other agents in our framework.


\subsection{Analyzer Agent}
Authors may not always disclose every limitation of their work, either due to oversight or to improve the chances of publication. Motivated by this, we introduce the Analyzer Agent, which is designed to infer implicit limitations. These are weaknesses that are not explicitly stated but are evident from the study’s design, statistical analysis, or scope of findings. Acting as a methodological auditor, the agent reviews the entire paper to detect issues such as unaddressed confounding variables, inadequate statistical power, or missing corrections for multiple comparisons. For each weakness it identifies, the agent generates a clear, evidence-based critique explaining how the issue might undermine the study's validity, reliability, or generalizability. This process allows it to supplement the author-stated limitations with deeper, more critical analysis. This inferential process follows our “chain of limitations” approach. The agent first examines the paper's methodology to identify a potential gap. It then formulates an argument for why the gap constitutes a limitation, grounding its reasoning in the text. This critique is then re-evaluated for accuracy before the agent moves on, ensuring detailed coverage across all key aspects of the study.


\subsection{Reviewer Agent} 
Top-tier conferences like ICLR \cite{b3} and AAAI \cite{b4} have begun experimenting with LLMs in the peer-review process, tasking them with generating pros and cons for submitted papers. Inspired by this trend, we introduce the Reviewer Agent, which generates limitations from the perspective of an external peer reviewer. Acting as an independent reviewer, this agent evaluates the manuscript for issues related to reproducibility, transparency, and ethical considerations. It assesses whether methods are described in sufficient detail, if data and code availability are addressed, and if ethical disclosures meet community standards. Drawing on best practices from peer review, the agent surfaces the types of limitations a human reviewer would typically raise, such as missing randomization details, or an insufficient data-sharing policy. This process follows our ``chain of limitations" approach. The agent adopts a structured review lens, focusing on a key area like reproducibility. It then identifies a potential shortcoming, justifies it with evidence from the text, and synthesizes its findings into a formal critique. Finally, the agent iteratively re-analyzes the paper to ensure all critical issues within its scope are captured. 

\subsection{Citation Agent} 
The content of a paper alone is often insufficient to capture all of its limitations. Authors may overlook certain shortcomings, or the broader impact of their work may only become apparent when examined in the context of related research. In this way, limitations flow not only from what is written in the input paper but also from how it interacts with the surrounding body of work. To address this, we introduce the Citation Agent, which leverages relevant literature to generate more meaningful limitations of the input paper. The agent collects two categories of related texts: (a) \textit{Cited In}: papers referenced by the input paper, and (b) \textit{Cited By}: papers that later cite the input paper. By analyzing both perspectives, the Citation Agent extracts contextually grounded insights that may not be evident within the paper itself. By incorporating input paper and cited works, the citation agent generates limitations that align with the input paper but are enriched by evidence and critiques found in these related works. This allows the system to move beyond the narrow scope of self-reported or inferred limitations, producing a more thorough and realistic assessment of the paper’s weaknesses.
The Citation Agent operates using an autonomous toolchain (ScienceParse, OpenAlex, and Selenium) to collect and process these related works. To identify Cited In papers, it uses ScienceParse to extract metadata from the input paper’s reference section. For Cited By papers, it queries the OpenAlex API. Each identified paper is then cross-referenced with the arXiv repository, and when a match is found, the corresponding preprint is downloaded and parsed with ScienceParse. This pipeline allows the agent to ground its generated limitations in a wide body of relevant literature.

\textbf{Retrieval Augmented Generation:}  To power the Citation Agent's ability to draw insights from the surrounding literature, we implemented a sophisticated Retrieval-Augmented Generation (RAG) system. Our goal was not just to find topically similar text, but to design a multi-stage process that identifies passages from cited works that are highly relevant to the specific context of the input paper, $P_i$. 
The process begins by constructing a unique vector database for each input paper, populated with the text of its Cited In and Cited By papers. Each cited paper in this database is divided into sections, with each section stored as an individual chunk. The total number of chunks for a given paper $P_i$ is defined as: 
\[
C_{total}(P_i) = (X_{in} \times A_{in}) + (X_{by} \times A_{by})
\]  
where $C_{total}(P_i)$ denotes the total number of chunks for input paper $P_i$, $X_{in}$ is the number of papers in the Cited In list, $A_{in}$ is the average number of sections per Cited In paper, $X_{by}$ is the number of papers in the Cited By list, and $A_{by}$ is the average number of sections per Cited By paper.
Our retrieval and re-ranking pipeline then unfolds in two stages. First, we use a hybrid retrieval approach that combines sparse (BM25) and dense (FAISS) methods with equal weighting to rank all chunks and select the top 20 candidates from $C_{total}(P_i)$. Second, these candidates are passed to an LLM-based re-ranker, which evaluates their direct relevance to the input paper $P_i$. Only chunks receiving a relevance score of 8 or higher are retained. These highly relevant chunks, along with the input paper itself, are then provided to the Citation Agent’s LLM generator. This rigorous pipeline ensures that the generated limitations are not only meaningful but also tightly aligned with the context of the input paper.

\subsection{Judge Agent} 
\label{sec:judge-self-feedback-agent}

To ensure the quality and reliability of our multi-agent system, we introduce the Judge Agent. This agent serves as an automated "quality auditor" with two complementary roles: performing a reference-free assessment of intrinsic quality and a reference-based validation against the ground truth. First, the Judge Agent conducts a reference-free evaluation of each worker agent's output. It scores the generated limitations across several dimensions—including depth, originality, actionability, and topic coverage—to provide numeric scores and qualitative feedback on each agent's contribution. Second, the agent validates the generated limitations against the ground truth for faithfulness, soundness, and importance. This validation follows a reflective two-stage process: the LLM first evaluates the outputs using only the source paper, then revisits its judgments in light of the ground truth to refine its assessment. This allows us to analyze how the agent's evaluation changes when provided with a reference, offering deeper insights into its reasoning.


\subsection{Self-feedback Agent} 
Self-feedback is critical in a multi-agent system as it allows each component to iteratively improve its outputs before the final synthesis. Our Self-Feedback Agent is responsible for managing this refinement process.
When the Judge Agent's evaluation of a worker agent's output falls below a predefined quality threshold (e.g., an average score of 8/10), the Self-Feedback Agent is activated. It takes the Judge's detailed feedback—which highlights strengths, pinpoints deficiencies, and offers actionable suggestions—and combines it with the original paper and prompt. This consolidated information is then used to trigger a re-generation request to the underperforming worker agent.
This feedback-driven loop was repeated twice in our experiments to study the impact of refinement. After each regeneration, the Judge Agent re-evaluates the updated output, and the new feedback can be used for a subsequent cycle. By enforcing this iterative validation, the Self-Feedback Agent acts as a quality gate, ensuring that only well-substantiated, novel, and useful limitations progress to the final synthesis by the Master Agent.

\subsection{Master Agent} 
After the worker agents have generated and refined their outputs, the Master Agent performs the final, crucial step of synthesis. Because multiple worker agents can produce overlapping or fragmented outputs, this integration is essential for producing a cohesive and professional result.
The Master Agent’s primary role is to consolidate all validated limitations from the Extractor, Analyzer, Reviewer, and Citation Agents into a single, unified list. To achieve this, it clusters similar statements using thematic similarity, merges redundant items while retaining the most exhaustive phrasing, and carefully preserves provenance tags (e.g., “Author-stated,” “Inferred,” or “Peer-review-derived”) for traceability.
The agent's workflow begins by receiving the validated outputs from all worker agents. It first resolves discrepancies and then groups related limitations before merging them into a single, prioritized statement. For each consolidated item, the Master Agent also generates a succinct justification—citing the relevant article section or supporting literature—to ensure clarity and scientific rigor. Finally, it performs a completeness check, iterating on the synthesis if necessary to ensure the final output is thorough. 

\begin{figure}[ht]
    \centering
    \begin{tikzpicture}
        \node[draw, rectangle, rounded corners, fill=yellow!20, text width=0.51\textwidth, inner sep=5pt, scale=0.9] (box) {
            \begin{minipage}{\textwidth}
                \small                
                \textbf{Extractor Agent} = 
                You are an expert in scientific literature analysis. 
                Extract all explicitly stated limitations as mentioned by the authors. Focus on sections such as the Discussion, Conclusion. 
                ....... 
               
                \vspace{2mm}\hrule\vspace{2mm}
                
                \textbf{Analyzer Agent} = You are a critical scientific reviewer with expertise in research methodology and analysis. Your task is to analyze the provided scientific article and identify potential limitations that are not explicitly stated by the authors. Focus on aspects 
                such as study design, sample size, data collection... 
                
                
                \vspace{2mm}\hrule\vspace{2mm}
                
                \textbf{Reviewer Agent} = You are an expert in open peer review with a focus on transparent and critical evaluation of scientific research. Your task is to review the provided scientific article from the perspective of an external peer reviewer. Identify potential limitations 
                that might be raised in an open review process, considering common critiques such as reproducibility, transparency, 
                generalizability....
                
                \vspace{2mm}\hrule\vspace{2mm}

                 \textbf{Citation Agent} = Your goal is to identify and articulate limitations for the current paper, focusing specifically on shortcomings that align with or are relevant to the findings, methods, or scope of the cited papers. Prioritize limitations that could be useful for strengthening the current paper’s discussion or future research. Ensure the limitations are concise, scientifically grounded, and directly tied to the context provided by the cited papers ...
               
                \vspace{2mm}\hrule\vspace{2mm}

                \textbf{Master Agent} = You are a **Master Coordinator**. 
                Your goals are to:
                1. Combine all limitations into a cohesive list, removing redundancies.
                2. Ensure each limitation is clearly stated, scientifically valid, and aligned with the article’s content.
                3. Prioritize limitations explicitly mentioned by the authors, supplementing them with inferred or peer-review-based limitations only if they add value..... 
            \end{minipage}
        };
    \end{tikzpicture}
    \caption{Example prompt excerpts for the Extractor, Analyzer, Reviewer, Citation, and Master Agents.}
    \label{fig:prompt-llm-agents}
   \vspace{-12pt} 
\end{figure}

\section{Evaluation} 
Recognizing that traditional NLP metrics alone are insufficient and that using an LLM-as-Judge to holistically compare the entire generated text against the full ground truth can miss nuanced matches and produce inconsistent scores, we designed a more granular, two-pronged evaluation framework. This framework assesses both the quantity and quality of the generated limitations.

\subsection{Quantity: Ground Truth Coverage} 
\label{sec:cov}
A precision-oriented metric, which measures the proportion (number) of generated limitations that match the ground truth, is unreliable. The core issue is that any ground truth for scientific limitations is inherently incomplete, as valid shortcomings can be documented in subsequent Cited By papers or other sources not included in our dataset. Consequently, a generated limitation that doesn't match our ground truth is not necessarily a hallucination but could be a valid, novel insight. Calculating precision against our non-exhaustive ground truth would unfairly penalize the model for such discoveries.
Therefore, our quantitative evaluation focuses on a more robust, recall-oriented metric called Ground Truth Coverage \cite{b1}. This metric calculates the percentage of ground truth limitations that are successfully matched with a corresponding model-generated limitation. Each one-to-one match is determined using our point-wise LLM-as-Judge methodology. The final score is produced by calculating the coverage for each paper individually and then averaging these scores across the entire corpus and presenting it as a percentage.
To assess the reliability of the 'LLM as a Judge' framework for similarity detection, we conducted a validation study on a random sample of 100 pairs. Two independent human annotators evaluated these samples, and we compared their decisions against the LLM's outputs. The results showed a high degree of consistency between the model and human judgment, with agreement scores of 0.98 for Annotator 1 and 0.95 for Annotator 2.  

\subsubsection{\textbf{Setup and Notation}}
Let our corpus of papers be denoted by $P = \{p_1, p_2, \ldots, p_N\}$, where $N$ is the total number of papers. For each paper $p_i \in P$: Let $A_i = \{a_{i1}, a_{i2}, \ldots, a_{im_i}\}$ be the set of $m_i$ ground truth limitations, extracted by our LLM-based extractor. Let $B_i = \{b_{i1}, b_{i2}, \ldots, b_{ij_i}\}$ be the set of $j_i$ limitations produced by our generation model.

\subsubsection{\textbf{Pairwise Similarity Judgment}}
For each paper $p_i$, we create all possible pairs $(a_{ik}, b_{il})$ between a ground truth limitation $a_{ik} \in A_i$ and a generated limitation $b_{il} \in B_i$. Each pair is evaluated by an LLM serving as a judge. We define a binary similarity function, $S(a, b)$, which represents the judge's decision:

\begin{equation}
\label{eq:similarity_func}
S(a_{ik}, b_{il}) = 
\begin{cases} 
    1 & \text{if } a_{ik} \text{ and } b_{il} \text{ are judged as similar} \\ 
    0 & \text{otherwise} 
\end{cases}
\end{equation}

\subsubsection{\textbf{Paper-Level Ground Truth Coverage Score}}

Our goal is to determine if each ground truth limitation $a_{ik}$ is ``covered'' by at least one of the generated limitations in the set $B_i$. We define a match score, $M(a_{ik})$, for a single ground truth limitation as 1 if it matches any generated limitation, and 0 otherwise. This can be expressed using the max function over all pairwise similarity scores for that ground truth item:

\begin{equation}
\label{eq:match_score}
M(a_{ik}) = \max_{l=1}^{j_i} S(a_{ik}, b_{il})
\end{equation}

The overall score for a single paper $p_i$, which we term \textbf{Ground Truth Coverage ($\text{C}_{GT}$)}, is the average of these match scores across all ground truth limitations for that paper. This is effectively the proportion of ground truth limitations that were successfully captured by the model.

\begin{equation}
\label{eq:paper_coverage}
\text{C}_{GT}(p_i) = \frac{1}{m_i} \sum_{k=1}^{m_i} M(a_{ik}) = \frac{1}{m_i} \sum_{k=1}^{m_i} \left( \max_{l=1}^{j_i} S(a_{ik}, b_{il}) \right)
\end{equation}

\subsubsection{\textbf{Overall Performance Metric}}

The final performance metric is the mean of the Ground Truth Coverage scores across all $N$ papers, which is then reported as a percentage.

\begin{equation}
\label{eq:mean_coverage}
\text{Mean C}_{GT} = \frac{1}{N} \sum_{i=1}^{N} \text{C}_{GT}(p_i)
\end{equation}

\textbf{Coverage as a Full} 
\label{sec:cov-as-full}
As a baseline to our primary point-wise methodology, we also experimented with a more holistic approach, which we term Full Text Coverage. In this evaluation, we provided the LLM-as-Judge with the entire set of ground truth limitations and the full set of model-generated limitations for a given paper, $P_i$, and prompted it for a single, overall coverage score.




\subsection{Quality: Similarity of Matched Pairs} 
While our primary evaluation metric is Coverage (a measure of quantity), we also assess the Quality of matched pairs as a secondary objective. For each pair of matched ground truth and generated limitations, ($(a_{ik}, b_{il})$), identified during the coverage step, we quantify their textual and semantic similarity. This is done using a suite of traditional NLP-based metrics, including ROUGE-L, Cosine Similarity, and Jaccard Similarity.
We treat these quality scores as supplementary because, while useful, these traditional metrics are fundamentally limited. They prioritize surface-level similarity over true comprehension; lexical-based metrics like ROUGE and Jaccard Similarity penalize valid paraphrasing. Furthermore, semantic measures like Cosine Similarity can fail on logical contradictions. For example, a generated limitation can be very close to a ground truth limitation in the embedding space while directly contradicting its meaning in reality (e.g., ``the model is effective" vs. ``the model is ineffective"). Consequently, these tools cannot reliably measure the most essential qualities of the generated text, reinforcing our decision to focus on the more robust Coverage metric.

\begin{table*}[htbp]
  \centering
  \footnotesize
  \begin{tabular}{@{}l l c c c c c c@{}}
    \toprule
    \textbf{Model} & \textbf{Input} & \textbf{Agents} & \textbf{C\textsubscript{GT}} & \textbf{R-L} & \textbf{BLEU} & \textbf{CS} & \textbf{JS} \\  
    \midrule
    Llama 3 8B(zs) \cite{b1} & IP (All) & No Agents & 62.04 &  14.92 & 0.37 & 29.99 & 8.73 \\  
    Llama 3 8B (Proposed)  &  IP (All) + CP & 3 Agents  & \textbf{66.45} & 13.27 & 0.19 & 28.64 & 6.89 \\ 
    Llama 3 8B & IP (All) + CP & 3 Agents + Feedback & 53.83 & 14.02 & 0.2 & 29.67 & 7.79 \\ 
    Llama 3 8B & IP (All) + CP & 4 Agents (Exp 1) & 50.7 & 15.12 & \textbf{0.79} & 30.80 & 8.49 \\
    Llama 3 8B & IP (All) + CP & 4 Agents (Exp 2) & 56.12 & 15.11 & 0.73 & 31.27 & 8.57 \\
    Llama 3 8B & IP (All) + CP & 3 Agents + zs & 58.47 & 14.22 & 0.48 & 29.51 & 8.03 \\
    Llama 3 8B & IP (All) + CP & 9 Agents & 49.81 &  \textbf{16.05} & 0.69 & \textbf{31.52} & \textbf{8.89} \\ \midrule
    GPT 4om(zs) \cite{b1} & IP (3 sec) & No Agents & 45.69 &  17.07 & 0.4 & 35.42 & 9.12 \\ 
    GPT 4om(zs) \cite{b1} & IP (All) & No Agents & 49.43 & \textbf{18.83} & 0.6 & \textbf{38.87} & 10.38 \\  
    GPT 4om  &  IP (All) & 3 Agents & 57.85 & 17.43 & 0.55 & 31.58 & 9.51 \\ 
    GPT 4om (Proposed) &  IP (All) + CP & 4 Agents & \textbf{64.94} & 17.93 & \textbf{1.18} & 32.60 & \textbf{10.98} \\ 
    \bottomrule
  \end{tabular} 
  \vspace{5pt}
  \caption{Performance of Limitation generation with different models. Here, 3 Agents means Extractor, Analyzer, and Reviewer. 4 Agents means 3 Agents + Citation. Here, Citation agent uses only cited papers (Exp 1), Citation agent uses input + cited papers (Exp 2). Here, All, 3 sec, zs, IP, CP, C\textsubscript{GT}, BS, R-L, CS, JS denotes all sections excluding ground truth, top 3 sections, zero shot, Input Paper, Cited Papers, Coverage of Ground Truth, Rouge-L, Cosine Similarity, and Jaccard Similarity.}
  \label{tab:llm-agents} 
  \vspace{-12pt}
\end{table*} 
\section{Experimental Setup} 
In our framework, the ``worker agents" are the Extractor, Analyzer, Reviewer, and Citation agents, while the Judge and Self-Feedback agents serve as meta-agents for evaluation and refinement. We experimented with several models for our worker and master agents—including Llama 3 8B, GPT-4o mini, DeepSeek R1 Qwen Distil, and Gemini—ensuring the same LLM was used for all roles within a given experimental run. For all utility functions—including the initial data extraction, the Judge Agent, and the final pointwise evaluation—we exclusively used GPT-4o mini due to its robust performance. For all agents, the input was the full paper with author limitations removed, while the Citation Agent's input was expanded with its citation network (Cited In/By papers). The ground truth for all evaluations combined author-stated limitations with critiques sourced from OpenReview.

To optimize our framework, we identified the top three sections most relevant for generating limitations. By measuring cosine similarity with the dedicated Limitations section across our dataset, we determined that the Abstract, Introduction, and Results \& Experiments sections had the highest correlation. To accommodate the context window of Llama 3 8B (8,192 tokens), any input exceeding this limit was truncated.
Beyond our core system, we experimented with several additional agents to explore other avenues for limitation discovery. An Image Agent analyzed figures and captions using PDFFigures 2.0 and a vision-language model to identify visually-based limitations. A Graph Agent constructed a knowledge graph from the paper's text using Stanford CoreNLP to infer limitations from structured relationships. We also tested numerous other specialized agents (e.g., Methodology, Data, Reproducibility), but found them less effective in producing generalizable limitations across diverse papers. Moreover, the activation threshold for the Self-Feedback Agent's refinement loop was set to a score of 8/10. 
As a cost-effective alternative to our primary LLM-reranker, we experimented with a gap-based retrieval method in RAG. This approach ranks cited papers by their cosine similarity to the input paper and then selects the top group by identifying the largest drop-off (the "gap") in similarity scores between consecutive papers. While computationally inexpensive, this method resulted in slightly lower performance compared to our proposed hybrid retrieval and LLM-reranker pipeline. 
We also tested a dual-stage RAG approach, where retrieved chunks were partitioned into "primary" (high-relevance) and "secondary" (lower-relevance) sets based on a similarity threshold. This tiered method proved unsuccessful, as it disrupted the output's coherence and slightly lowered performance compared to our main single-stage pipeline.

\begin{table}[htbp]
  \centering
  \small
    \begin{tabular}{c c c}
    \toprule
     \textbf{Metrics} & \textbf{w/o LLM's feed} & \textbf{w LLM's feed}  \\ 
    \midrule 
    C\textsubscript{GT} & \textbf{66.45} & 53.83 \\
    C\textsubscript{LLM} & 36.59 & \textbf{44.77} \\  
    \bottomrule
  \end{tabular} 
  \vspace{3pt}
    \caption{Comparison of performance without and with LLM feedback using Llama 3 8B with 3 Agents. Here C\textsubscript{GT} and C\textsubscript{LLM} depicts Ground Truth Coverage and LLM Generated Text Coverage}

  \label{tab:with-without-feedback}
  \vspace{-7pt}
\end{table}

\begin{table}[htbp]
  \centering
  \small
    \begin{tabular}{c c c c}
    \toprule
     \textbf{Model} & \textbf{Agents} & \textbf{Input} & \textbf{Cov Full}  \\ 
    \midrule 
    Llama 3 8B (zs) & No Agents & IP & 20.62 \\
    Llama 3 8B & 3 Agents & IP & \textbf{30.06} \\
    Llama 3 8B & 4 Agents & IP + CP & 29.01 \\
    \bottomrule
  \end{tabular} 
  \vspace{3pt}
    \caption{Performance in Zero Shot (zs) and LLM Agents using 3 and 4 Agents}
  \label{tab:various-agents-full-cov}
  \vspace{-12pt}
\end{table}

\section{Experiments and Results} 
\subsection{PoitnWise Coverage Evaluation} 

We evaluated our framework using our primary metric, the point-wise Ground Truth Coverage (Cov\textsubscript{GT}). Our findings, summarized in Table \ref{tab:llm-agents}, reveal that the optimal agent configuration is highly dependent on the base model's capabilities.
For the smaller Llama 3 8B model, the 3-Agent configuration (Extractor, Analyzer, Reviewer) proved optimal. This setup achieved a Coverage score of 66.45\%, a +4.41 point improvement over the zero-shot baseline (62.04\%). Here, the input of each agent contains the input paper with cited papers. This success is due to the structured workflow, which assigns focused roles that reduce the model's cognitive load. For the more powerful GPT-4o mini, the 4-Agent configuration, which adds the Citation Agent, achieved the best results, yielding a substantial +15.51 point improvement over its zero-shot baseline and a +7.09 point boost over the 3-Agent version. Moreover, our optimized 3-Agent Llama 3 8B model not only surpasses the zero-shot performance of the much larger GPT-4o mini by +17.02 points but also slightly outperforms its optimal 4-Agent configuration by +1.51 points, all while offering a significantly more cost-effective solution. To further validate these results, we conducted a secondary, holistic coverage evaluation for the Llama 3 8B model (Table~\ref{tab:various-agents-full-cov}). This simpler metric corroborated our primary findings: the 3-agent approach again improved the score over the baseline (by +9.44 points), while adding the Citation Agent was once more detrimental, causing a -1.05 point drop.
Both evaluation methods confirm our key insight: a smaller model like Llama 3 8B performs best with a focused 3-agent setup because it struggles to effectively process the diverse text from cited papers as a citation agent. Conversely, a stronger model like GPT-4o mini successfully leverages this additional context to generate more in-depth limitations and improve coverage.
Our experiments with additional models highlighted the importance of a model's ability to follow complex instructions. Gemini 1.5 Flash struggled in the Extractor Agent role, failing to reliably identify relevant text, which in turn led to poor final outputs from the Master Agent. Similarly, the DeepSeek-R1-Distill-Qwen-7B model showed an incompatibility with our framework; our extensive, detailed prompts caused its worker agents to generate noisy and low-quality limitations. In contrast, the power of our structured agent-based approach was demonstrated with the much smaller Llama 3 1B model. While this model was unable to generate any limitations in a zero-shot setting, our framework successfully guided it to produce coherent outputs. This suggests that our approach not only enhances the performance of capable models but can also enable smaller models to tackle complex tasks.

\begin{table}[htbp]
  \centering
  \footnotesize
  \begin{tabular}{p{1.08cm} p{2.6cm} p{0.7cm} p{0.7cm} p{0.7cm}}
    \toprule
    \textbf{Agent} & \textbf{Metrics} & \textbf{w/o Feed} & \textbf{Feed. once} & \textbf{Feed. twice} \\  
    \midrule
    Extractor & Depth    &         6.17 & 6.82 & 
    \textbf{7.46}  \\
    Extractor & Originality &      5.33 & 6.03 & \textbf{6.48} \\
    Extractor & Actionability &      7.13 & 7.73 & \textbf{8.30} \\ 
    Extractor & Topic Coverage &   7.25 & 7.32 & \textbf{7.78} \\ 
    Analyzer  & Depth           &  7.99 & \textbf{8.14} & 7.71 \\ 
    Analyzer & Originality &        7.37 & \textbf{7.90} & 7.37 \\ 
    Analyzer & Actionability &      8.63 & \textbf{8.74} & 8.68 \\
    Analyzer & Topic Coverage &     \textbf{8.12} & 8.05 & 7.88 \\ 
    Reviewer & Depth &             8.11 & \textbf{8.47} & 7.92 \\ 
    Reviewer & Originality &        7.25 & \textbf{7.48} & 6.92 \\ 
    Reviewer & Actionability &       8.27 & \textbf{8.81} & 8.76 \\ 
    Reviewer & Topic Coverage &      8.33 & \textbf{8.63} & 7.89 \\ 
    Citation & Depth &               7.83 & \textbf{8.67} & 8.56 \\ 
    Citation & Originality &        6.96 & \textbf{8.08} & 7.94 \\ 
    Citation & Actionability &       8.24 & \textbf{8.42} & 8.38 \\ 
    Citation & Topic Coverage &      8.30 & \textbf{8.85} & 8.79 \\ 
    
    \bottomrule
  \end{tabular} 
  \vspace{3pt}
  \caption{Performance of various metrics of without (w/o) feedback, feedback once, and twice using GPT 4o mini}
  \label{tab:coverage-various-llm-agents-gpt} 
  \vspace{-9pt}
\end{table} 

\begin{table}[htbp]
  \centering
  \small
    \begin{tabular}{p{1.05cm} >{\centering\arraybackslash}p{0.75cm} >{\raggedleft\arraybackslash}p{1.2cm} >{\raggedleft\arraybackslash}p{1.2cm} >{\raggedleft\arraybackslash}p{1.2cm}}
    \toprule
    \textbf{Agent type} & \textbf{Iteration} & \textbf{Faith.} & \textbf{Sound.} & \textbf{Import.} \\
    \midrule
    Extractor & 1 & 3.258 & \textbf{2.662} & \textbf{3.289} \\
    Extractor & 2 & \textbf{3.369} & 2.633 & 3.212 \\ \midrule
    Analyzer  & 1 & 3.169 & \textbf{2.720} & 3.274 \\
    Analyzer  & 2 & \textbf{3.329} & 2.705 & \textbf{3.311} \\ \midrule
    Reviewer  & 1 & \textbf{3.215} & 2.680 & \textbf{3.289} \\
    Reviewer  & 2 & 3.188 & \textbf{2.668} & 3.234 \\
    \bottomrule
  \end{tabular}
  \vspace{3pt}
  \caption{Agent-wise evaluation across two iterations (with or without feedback) using Llama 3 8B. Feedback added in 2nd iteration. Metrics are Faithfulness, Soundness, and Importance.}
  \label{tab:llama-feedback-llm-agents}
  \vspace{-11pt}
\end{table}

\subsection{Incorporating Feedback} 
We investigated the impact of a self-feedback loop, where critiques from the LLM-as-a-Judge were used to prompt a second round of limitation generation. Our findings reveal a critical trade-off (Table \ref{tab:with-without-feedback}) in the Llama 3 8B model: while feedback improved the coverage of the LLM-generated limitations (C \textsubscript{LLM}) (+8.18) (precision), it significantly decreased our primary metric, Ground Truth Coverage (Recall) (-12.62). This occurs because the feedback constrains the model’s reasoning, causing it to focus on refining prior points rather than exploring new ones. The resulting limitations are more polished and informative, but their reduced diversity and quantity lower the overall recall of the full ground truth. Moreover, as shown in Table \ref{tab:llama-feedback-llm-agents}, incorporating feedback did not consistently enhance the quality metrics (soundness, importance) for individual agents; in several cases, performance deteriorated. This effect was particularly pronounced in the smaller Llama 3 8B model, which suggests that smaller models may lack the capacity to effectively integrate and act upon nuanced feedback for regeneration.
In contrast, the more capable GPT-4o mini demonstrated a pattern of initial improvement followed by degradation (Table \ref{tab:coverage-various-llm-agents-gpt}). One iteration of feedback led to broad improvements across nearly all agents and quality metrics (Depth, Originality, Actionability, etc.). However, a second iteration proved counterproductive, causing performance to drop for most agents. This indicates that while a stronger model can effectively leverage a single round of feedback, repeated refinement introduces diminishing returns and can even degrade the output.

\subsection{Quality Evaluation} 

Our quality evaluation using NLP-based similarity metrics revealed a complex trade-off between the quality of individual matches and overall coverage (Table \ref{tab:llm-agents}). For the Llama 3 8B model, a 9-agent configuration achieved the highest quality scores, but coverage drops (-16.64) by a big margin. This setup produced fewer but more detailed limitations, resulting in higher-quality textual matches when they did align with the ground truth, but at the cost of missing many other points. For GPT-4o mini, our proposed 4-agent setup excelled on metrics like BLEU and Jaccard Similarity (+0.78, +1.86 than zero-shot). However, the zero-shot baseline performed best on ROUGE-L and Cosine similarity because the 4-agent model's use of cited papers introduced relevant, external information that did not textually overlap with our ground truth, thereby lowering certain similarity scores.
 


\begin{table}[htbp]
  \centering
  \footnotesize
  \begin{tabular}{p{3.50cm} p{0.9cm} p{0.7cm} p{0.7cm} p{0.7cm}}
    \toprule
    \textbf{LLM Agents} & \textbf{Eval Type} & \textbf{Faith} & \textbf{Sound} & \textbf{Imp} \\  
    \midrule
    Ext & Type 1 & 4.00 & 3.00 & 3.98 \\ 
    Ext & Type 2 & 2.82 & 2.30 & 2.53 \\ 
    \midrule
    Ext + Analy & Type 1 & 4.00 & 3.00 & 3.99 \\ 
    Ext + Analy & Type 2 & 3.22 & 3.30 & 3.15 \\ 
    \midrule
    Ext + Analy + Rev & Type 1 & 4.00 & 3.00 & 3.98 \\ 
    Ext + Analy + Rev & Type 2 & 3.24 & 3.61 & 3.26 \\ 
    \midrule
    Ext + Analy + Rev + Cit & Type 1 & 4.00 & 3.00 & 3.98 \\ 
    Ext + Analy + Rev + Cit & Type 2 & \textbf{3.36} & \textbf{3.71} & \textbf{3.40} \\ 
    \bottomrule
  \end{tabular} 
  \vspace{3pt}
  \caption{Ablation study on Multi LLM agents settings. Here, GPT-4o mini is used as an agent and evaluator. Here, Type 1 is `Input with LLM-Generated Limitation' and Type 2 is `Revised Ground Truth with LLM-Generated Limitation'. LLM Agents are Ext (Extractor), Analy (Analyzer), Rev (Reviewer), and Cit (Citation). Metrics are Faith (Faithfulness), Sound (Soundness), and Imp (Importance)}
  \label{tab:llm-agents-ablation-faith-sound-importance} 
  \vspace{-10pt}
\end{table}


\begin{table}[htbp]
  \centering
  \footnotesize 
  \begin{tabular}{p{1.98cm} p{4.6cm} p{0.7cm}}
    \toprule
    \textbf{Model} & \textbf{Agent type} & \textbf{C\textsubscript{GT}} \\  
    \midrule
    GPT 4o mini & Methodology, Assumption, data, and master agent & 40 \\  \midrule
    GPT 4o mini & Experiment, clarity, impact, and master agent & 38.63  \\  \midrule 
    GPT 4o mini & Extractor &  43.39 \\  
    GPT 4o mini & Extractor, Analyzer, and master agent & 30.78 \\  
    GPT 4o mini & Extractor, Analyzer, Reviewer, and Master agent
    & 57.85  \\ 
    GPT 4o mini & Extractor, Analyzer, Reviewer, Graph, and Master agent
    & 14.20 \\ 
    \textbf{Llama 3 8B + RAG} & Extractor, Analyzer, Reviewer, Citation, Theory, Baselines, Robustness, Reproducibility, and Presentation & 49.81  \\ 
    \textbf{GPT 4o mini + RAG} & \textbf{Extractor, Analyzer, Reviewer, citation, and master agent} & \textbf{64.94}  \\ 
    \bottomrule
  \end{tabular}
  \vspace{4pt}
  \caption{Coverage of Ground Truth (C\textsubscript{GT}) using various types of LLM Agents settings as an ablation study.}
  \label{tab:various-llm-agents} 
  \vspace{-12pt}
\end{table} 

\section{Ablation Study}  


\textbf{Impact of Citation Agent Context (Llama 3 8B):} Our study on the Citation Agent found that the input paper's context to effectively align limitations from external works, making its generated limitations more relevant. Here, providing the agent with both the input paper and cited papers increased (Expt 2 Table \ref{tab:llm-agents}) in  Coverage by +5.42 points over using cited papers alone (Expt 1). 

\textbf{Impact of Agent Quantity (Llama 3 8B):}
We also investigated the effect of the number of agents on the Llama 3 8B model. A naive combination of the 3-Agent output with the zero-shot baseline degraded performance by -3.57 points (Table \ref{tab:llm-agents}) than zero-shot alone. Pushing the architecture further to nine agents proved counterproductive, causing a significant performance drop of -12.23 points. This suggests that increasing the number of agents beyond a certain point introduces excessive noise that overwhelms their ability to produce coherent outputs.


\textbf{Impact of Input Granularity:} For the GPT-4o mini model, using the full text of the input paper rather than just the top three sections improved performance by Coverage by +3.74 points and quality metrics such as ROUGE-L (+1.76) (\ref{tab:llm-agents}).  
\textbf{Ablation on Core Agent Contributions} 
We conducted a sequential ablation study on our four core agents (using GPT-4o mini) to measure their incremental impact on both Coverage and qualitative metrics like Faithfulness, Soundness (Tables \ref{tab:various-llm-agents} \& \ref{tab:llm-agents-ablation-faith-sound-importance}).
The Extractor Agent alone set a modest baseline. Adding the Analyzer Agent created a key trade-off, decreasing Coverage (-12.61) while substantially increasing quality. The Reviewer Agent then acted as a crucial synthesizer, dramatically boosting both Coverage (+27.07) and quality. Finally, the Citation Agent provided a final improvement across all metrics (+7.09 on Coverage), confirming that the full 4-agent configuration is essential for balancing broad coverage with high-quality outputs.

\textbf{Comparison with Specialized Agent Roles:}  We experimented with alternative, more specialized agent configurations to validate our choice of the four core agents, but they consistently underperformed. Agents with narrow roles (e.g., Methodology, Data) dropped Coverage scores by up to 15 points, while a Graph Agent was particularly detrimental, causing a performance collapse of -50.74 points \ref{tab:various-llm-agents}. This failure stems from their narrow focus on specific limitation categories that are not present in all papers. In contrast, our proposed agents succeed because they are designed to be broad yet complementary, targeting different types of weaknesses. 

\textbf{Ablation on Chunk Enrichment}
We investigated the impact of chunk enrichment on retrieval. We used an LLM to generate a summary for each cited paper and then prepended this summary to every chunk (section) from that paper before indexing. This approach yielded a modest +1\% improvement in Coverage, suggesting that providing high-level context directly within each chunk can be beneficial.


\textbf{Importance of Reference:} We found our LLM-as-a-Judge is reliable only when provided with a ground truth reference. Without a reference, its scores for different agents were not meaningfully distinct (4,3,3.98); however, including the ground truth produced granular and reliable scores. (Table \ref{tab:llm-agents-ablation-faith-sound-importance}).


\section{Discussion} 
\textbf{RQ1:} Can LLM agents outperform zero-shot.... 

Yes, our findings confirm that a structured LLM agent framework consistently outperforms zero-shot prompting for both smaller and larger models. Our optimal 3-agent setup with Llama 3 8B and our 4-agent setup with GPT-4o mini both achieved substantial gains (+4.41, +15.51) in coverage over their respective zero-shot baselines.
We attribute this success to the agent-based design, which distributes complex reasoning into focused roles like extraction, analysis, and synthesis. Unlike a single, unstructured pass, this staged workflow mirrors the systematic process of human peer review, enabling a more extensive identification of a paper's limitations. 


\textbf{RQ2:} What is the optimal agent configuration...
Our findings show the optimal agent configuration depends on model capacity: a focused 3-agent setup was best for the smaller Llama 3 8B, while the more powerful GPT-4o mini benefited from a 4-agent setup that included a Citation Agent. Because a smaller model Llama 3 8B couldn't process the text from diverse sources as a citation agent. Moreover, a sequential ablation study on GPT-4o mini confirmed that the addition of each of our four core agents incrementally improved output quality. This success stems from our task-oriented agent design, which proved more flexible than the overly specialized agent roles that consistently underperformed (e.g., Methodology, Data). While this decomposition of tasks improves reasoning over a single zero-shot pass, we found that excessive decomposition (e.g., a 9-agent setup) is counterproductive, introducing noise that harms the breadth of the results even as it improves the detail of individual matches.



\textbf{RQ3:} How reliable are different evaluation strategies... 


Our analysis reveals a clear trade-off between measuring the breadth (the diversity of limitations captured) and the depth (the textual quality of each). We found that our point-wise Ground Truth Coverage is the most reliable metric for assessing breadth, as it faithfully signals how well a system identifies a wide range of distinct issues. In contrast, for measuring depth (as a secondary metric), we used traditional NLP-based metrics, which primarily capture surface-level similarity and struggle with valid paraphrasing. Moreover, the feedback-driven refinement directly manipulates this trade-off in Llama 3 8B, which failed to effectively leverage the feedback, generally improving depth at the expense of breadth. However, this focus on refinement constrained the models' generative scope, reducing diversity and causing a significant drop in our primary coverage metric. While the more powerful GPT-4o mini successfully used a single round of feedback to improve depth, a second round proved counterproductive, leading to diminishing returns and performance degradation. For both models, feedback produced limitations that were more polished and coherent. 

\section{Conclusion} 
In this work, we introduce an LLM agent-based framework designed to address the common underreporting of scientific limitations. To enable a robust analysis, we constructed a extensive ground truth by combining author-stated limitations with peer-review feedback and developed a granular point-wise evaluation method that surpasses traditional NLP metrics.
Our experiments show that this agent-based approach consistently outperforms zero-shot baselines, with the optimal agent configuration depending on the model's capacity. Furthermore, our analysis reveals a critical trade-off: feedback-driven refinement enhances the quality (depth) of limitations but reduces their variety (breadth). Ultimately, our findings demonstrate that agent-based decomposition is a powerful strategy for automatically generating a more complete and transparent view of a study's limitations.

\section{Limitations and Future Work} 
Our study has several limitations. First, our findings are based solely on the NeurIPS dataset and only two models (Llama 3 8B and GPT-4o mini), which restricts generalizability. Methodologically, we truncated long inputs for the Llama model, our human evaluation was confined to the initial extraction task, and our detailed prompts were incompatible with certain reasoning-oriented models like DeepSeek. Future work will address these points by expanding our evaluation to more domains and a wider range of open-source models. We also plan to implement chunking strategies to avoid input truncation and conduct a more comprehensive human evaluation.  



\section{Appendix} 

\begin{figure*}[htbp]
    \centering
    \begin{tikzpicture}
        \node[draw, rectangle, rounded corners, fill=yellow!20, text width=0.85\textwidth, inner sep=5pt] (box) {
            \begin{minipage}{\textwidth}
                \small
                \raggedright 
                \textbf{Prompt} = \texttt{\textquotesingle\textquotesingle\textquotesingle} \\
                You are an expert in scientific literature analysis. Your task is to carefully read the provided scientific article
and extract all explicitly stated limitations as mentioned by the authors. Focus on sections such as Discussion,Conclusion, or
Limitations. List each limitation verbatim, including direct quotes where possible, and provide a brief context (e.g., what aspect of
the study the limitation pertains to). Ensure accuracy and avoid inferring or adding limitations not explicitly stated. If no limitations
are mentioned, state this clearly.

Workflow:

Plan: Outline which sections (e.g., Discussion, Conclusion, Limitations) to analyze and identify tools (e.g., text extraction) to
access the article content. Justify the selection of sections based on their likelihood of containing limitation statements.

Reasoning: Let's think step by step to ensure thorough and accurate extraction of limitations:
Step 1: Identify all sections in the article that may contain limitations. For example, the Discussion often includes
limitations as authors reflect on their findings, while a dedicated Limitations section is explicit.

Step 2: Use text extraction tools to retrieve content from these sections. Verify that the content is complete and accurate. 

Step 3: Scan for explicit limitation statements, such as phrases like "a limitation of this study" or "we acknowledge that." Document why each statement qualifies as a limitation. 

Step 4: For each identified limitation, extract the verbatim quote (if available) and note the context (e.g., related to sample size,
methodology). 

Step 5: Check for completeness by reviewing other potential sections (e.g., Conclusion) to ensure no limitations are missed.
Analyze: Use tools to extract and verify the article's content, focusing on explicit limitation statements.
Cross-reference extracted
quotes with the original text to ensure accuracy.

Reflect: Verify that all relevant sections were checked and no limitations were missed. Consider whether any
section might have been
overlooked and re-evaluate if necessary.
Continue: Do not terminate until all explicitly stated limitations are identified or confirmed absent.

Output Format:
Bullet points listing each limitation.
For each: Verbatim quote (if available), context (e.g., aspect of the study), and section reference.
If none: "No limitations explicitly stated in the article."

Tool Use:
Use text extraction tools to access and verify article content.
Do not assume content; retrieve it directly from the provided article.

Chain of Thoughts:
During the Reasoning step, document the thought process explicitly. For example:
"I selected the Discussion section because authors often discuss study constraints there."
"I found the phrase 'a limitation of this study' in the Limitations section, indicating an explicit limitation."
"I checked the Conclusion section to ensure no additional limitations were mentioned, confirming completeness."
This narrative ensures transparency and justifies each decision in the extraction process.                \texttt{\textquotesingle\textquotesingle\textquotesingle}
            \end{minipage}
        };
    \end{tikzpicture}
    \caption{Extractor Agent}
    \label{fig:extractor-agent}
\end{figure*}

\begin{figure*}[htbp]
    \centering
    \begin{tikzpicture}
        \node[draw, rectangle, rounded corners, fill=yellow!20, text width=0.85\textwidth, inner sep=5pt] (box) {
            \begin{minipage}{\textwidth}
                \small
                \raggedright 
                \textbf{Prompt} = \texttt{\textquotesingle\textquotesingle\textquotesingle} \\
You are a critical scientific reviewer with expertise in research methodology, data analysis, and theoretical evaluation,
acting as an Analyzer to identify potential limitations in a scientific article that are not explicitly stated by the authors.
Your task is to deeply analyze the provided article, focusing on study design, sample size, data collection methods, statistical analysis,
scope of findings, underlying assumptions, and their interconnections. Infer limitations by identifying subtle methodological, analytical,
or theoretical gaps, providing a rigorous explanation of why each is a limitation and its impact on the study’s validity, reliability,
generalizability, or practical implications. Ensure inferences are grounded in the article’s content, avoid speculative assumptions,
and think critically to uncover nuanced or overlooked shortcomings.

Input:

Article: [Full text of the scientific article, including sections such as Introduction, Methodology, Dataset, Experiment and Results,
Discussion, Conclusion, etc.]

Instructions:
Deep Critical Analysis: Examine the article with a layered approach, exploring:
Study Design: Are there flaws in the design (e.g., lack of controls, non-random sampling)? How do these affect outcomes?

Sample Size: Is the sample size sufficient for the study’s claims? Are there biases in selection or representation?
Data Collection: Are methods robust, reproducible, or prone to bias (e.g., self-reporting, incomplete datasets)?
Statistical Analysis: Are methods appropriate? Are there unaddressed confounders, overfitting, or lack of robustness checks?
Scope and Generalizability: Are findings overly specific to the context? Are claims justified by the data?
Assumptions: What implicit or explicit assumptions underlie the study? Are they tested or potentially flawed?
Interconnections: How do limitations in one area (e.g., methodology) impact others (e.g., results or scope)?

Infer Limitations: Identify limitations not explicitly stated by the authors, focusing on gaps that could undermine the study’s
rigor or applicability. Consider secondary effects (e.g., how a small sample size might amplify biases in analysis).

Ground Inferences: Base all limitations on the article’s content, using textual evidence to support inferences. Avoid external
assumptions or speculation beyond what the text supports.

Evaluate Impact: For each limitation, assess its effect on validity (e.g., internal consistency), reliability (e.g., reproducibility),
generalizability (e.g., applicability to other contexts), or practical implications (e.g., real-world utility).

Handle Edge Cases:
If data descriptions are incomplete, note potential limitations in transparency or reproducibility.
If statistical methods are ambiguous, infer limitations in analytical rigor.
If the scope is unclear, evaluate whether claims overreach the evidence.

Ensure Comprehensive Coverage: Analyze all relevant sections (e.g., Methodology, Results, Discussion) and cross-check for interconnected
limitations. Consider alternative perspectives (e.g., what a different methodology might reveal).
Avoid Redundancy: Consolidate related limitations into a single, clear point, ensuring each limitation is unique and significant.

Workflow:

Plan: Identify key areas for analysis (e.g., study design, statistical methods, assumptions) and justify their selection based on
their potential to reveal limitations. Plan to use text analysis tools to extract and verify details.

Reasoning:

Step 1: Break down the article by section, extracting details on methodology, sample size, data collection, analysis, scope, and
assumptions.
Step 2: Ask critical questions for each area, e.g.:

Methodology: “Is the design robust? Are controls adequate?”
Sample Size: “Does the sample size support the claims? Are there selection biases?”
Analysis: “Are statistical methods justified? Are confounders addressed?”

Assumptions: “What assumptions are untested? How might they skew results?”
Step 3: Identify gaps or weaknesses, inferring limitations based on evidence in the text.
Step 4: Cross-check sections to uncover interconnected limitations (e.g., how data collection affects analysis).
Step 5: Document each limitation with a clear explanation and impact, supported by textual evidence.
Step 6: Re-evaluate for overlooked gaps, considering alternative perspectives or trade-offs (e.g., benefits vs. risks of the chosen
method).
Analyze: Use advanced text analysis tools (e.g., dependency parsing for methodological details, keyword extraction for assumptions)
to confirm content and identify gaps.
Reflect: Verify that inferred limitations are grounded, significant, and not redundant. Re-assess ambiguous areas (e.g., vague statistical
descriptions) to ensure thoroughness.
Continue: Iterate until all potential limitations are identified, ensuring a deep and comprehensive analysis.

Tool Use:
Use dependency parsing to analyze complex sentences in Methodology or Results for hidden gaps.

Chain of Thoughts: Document the reasoning process internally for transparency. For example:
“The Methodology uses a convenience sample, which may introduce selection bias, limiting generalizability.”
“The statistical analysis omits adjustment for confounders, potentially reducing validity; this is a limitation because it affects
causal inference.”
“The assumption of data homogeneity is untested, which could skew results if variability exists.”
“Cross-checked Results to confirm no robustness checks were performed, reinforcing the limitation.”

Output Format: The output must be in plain text format, consisting of bullet points listing each inferred limitation. For each
limitation, include:
Explanation: A detailed explanation of why this is a limitation, grounded in the article’s content, including textual evidence or
reasoning.
Impact: The effect on the study’s validity, reliability, generalizability, or practical implications. If no limitations are inferred,
output:
“No inferred limitations identified based on the article’s content.”
''
     
             \texttt{\textquotesingle\textquotesingle\textquotesingle}
            \end{minipage}
        };
    \end{tikzpicture}
    \caption{Analyzer Agent}
    \label{fig:analyzer-agent}
\end{figure*}

\begin{figure*}[htbp]
    \centering
    \begin{tikzpicture}
        \node[draw, rectangle, rounded corners, fill=yellow!20, text width=0.85\textwidth, inner sep=5pt] (box) {
            \begin{minipage}{\textwidth}
                \small
                \raggedright 
                \textbf{Prompt} = \texttt{\textquotesingle\textquotesingle\textquotesingle} \\
You are an expert in open peer review with a focus on transparent and critical evaluation of scientific research.
Your task is to review the provided scientific article from the perspective of an external peer reviewer.
Identify potential limitations
that might be raised in an open review process, considering common critiques such as reproducibility, transparency, generalizability,
or ethical considerations. Leverage insights from similar studies or common methodological issues in the field by searching the web or
X posts for context, if needed.

Workflow:
Plan: Identify areas for review (e.g., reproducibility, transparency, ethics) and plan searches for external context
(e.g., similar studies, methodological critiques). Justify the selection based on peer review standards.
Reasoning: Let's think step by step to identify peer-review limitations:

Step 1: Select key areas for review (e.g., reproducibility, ethics) based on common peer review critiques. 

Step 2: Use text analysis tools to extract relevant article details (e.g., methods, data reporting). 

Step 3: Identify potential limitations, such as lack of transparency in data or ethical concerns, and justify using article content. 

Step 4: Search web/X for external context (e.g., similar studies) to support limitations, rating source relevance (high, medium, low, none).

Step 5: Synthesize findings, ensuring limitations align with peer review standards and are supported by article or external context.
Analyze: Critically review the article, integrating external context to identify limitations.

Use tools to verify content and sources.

Reflect: Verify that limitations align with peer review standards and are supported by the article or external context.
Re-evaluate overlooked areas if necessary.

Continue: Iterate until all relevant peer-review limitations are identified.

Output Format:
Bullet points listing each limitation.
For each: Description, why it's a concern, and alignment with peer review standards.
Include citations for external sources in the format Source Name, if used.

Tool Use:
Use web/X search tools to find relevant literature or methodological critiques.
Use text analysis tools to verify article content.

Chain of Thoughts:
During the Reasoning step, document the thought process explicitly. For example:
"I selected reproducibility because peer reviewers often critique data availability."
"The article lacks a data sharing statement, which limits reproducibility."
"A web search revealed similar studies provide data openly, supporting this limitation."
This narrative ensures transparency and justifies each identified limitation.

Paper Content:

Please provide a critical review identifying the limitations and areas of concern in this research. Consider what a peer reviewer would highlight as weaknesses or areas needing improvement.

              \texttt{\textquotesingle\textquotesingle\textquotesingle}
            \end{minipage}
        };
    \end{tikzpicture}
    \caption{Reviewer Agent}
    \label{fig:reviewer-agent}
\end{figure*}

\begin{figure*}[htbp]
    \centering
    \begin{tikzpicture}
        \node[draw, rectangle, rounded corners, fill=yellow!20, text width=0.85\textwidth, inner sep=5pt] (box) {
            \begin{minipage}{\textwidth}
                \small
                \raggedright 
                \textbf{Prompt} = \texttt{\textquotesingle\textquotesingle\textquotesingle} \\
You are an expert scientific research assistant tasked with inferring potential limitations for an unspecified
current scientific article based solely on its cited papers.
You are given information from multiple cited papers, which are assumed to be referenced by the current article.
Your goal is to analyze these cited works and identify possible limitations that the current paper may have, by
comparing its presumed scope, methods, or results against the cited literature.
Because the input paper itself is not provided, you must reason from the cited papers alone, identifying what
gaps, stronger methods, broader coverage, or alternative results the cited works might expose in the hypothetical
current paper that cites them.

Objective:

Generate a list of scientifically grounded limitations that the current article might have, assuming it builds upon or is informed by the provided cited papers.

Each limitation should:

Be concise

Reference the relevant cited paper(s) by title

Clearly explain how the cited paper exposes a potential limitation

Be plausible and insightful based on common scientific reasoning

Workflow:
Plan:
Identify key insights, strengths, and scopes of the cited papers that could set a high bar or reveal blind spots
in a hypothetical citing article.

Reasoning: Let's think step by step to infer limitations:
Review each cited paper to extract its methodology, findings, and scope.
Ask: If a paper cited this work but did not adopt or address its insights, what limitation might arise?
Identify where the cited paper offers better methodology, broader scope, or contradicting findings.
Formulate each limitation as a plausible shortcoming of a hypothetical article that builds on—but possibly
underutilizes—these cited works.

Justify each limitation based on specific attributes of the cited paper (e.g., "more comprehensive dataset",
"stronger evaluation metric", etc.)

Analyze:
Develop a set of inferred limitations, each tied to specific cited paper(s) and grounded in logical comparison.

Reflect:
Ensure coverage of all relevant cited papers and validate that each limitation is scientifically plausible in
context.

Output Format:
Bullet points listing each limitation.
For each: Description, explanation, and reference to the cited paper(s) in the format Paper Title.

Tool Use (if applicable):

Use citation lookup tools or document content to extract accurate summaries.
Do not assume details about the input paper—focus only on drawing limitations based on differences, omissions,
or underuse of the cited works.

Chain of Thoughts:
During the Reasoning step, document the thought process explicitly. For example:
"I selected [Paper X] because it uses a more robust method than the current article."
"The current article's simpler method may limit accuracy compared to [Paper X]."
"I reviewed all cited papers to ensure no relevant gaps were missed."
This narrative ensures transparency and justifies each identified limitation.

Input paper: 

Cited Papers:

Please identify limitations that would be relevant for researchers who might cite this paper in future work.
Consider what limitations future authors might mention when discussing this paper's contribution to the field,
based on the cited papers context.

              \texttt{\textquotesingle\textquotesingle\textquotesingle}
            \end{minipage}
        };
    \end{tikzpicture}
    \caption{Citation Agent}
    \label{fig:citation-agent}
\end{figure*}

\begin{figure*}[htbp]
    \centering
    \begin{tikzpicture}
        \node[draw, rectangle, rounded corners, fill=yellow!20, text width=0.85\textwidth, inner sep=5pt] (box) {
            \begin{minipage}{\textwidth}
                \small
                \raggedright 
                \textbf{Prompt} = \texttt{\textquotesingle\textquotesingle\textquotesingle} \\
You are a **Master Coordinator**, an expert in scientific communication and synthesis. Your task is to integrate
limitations provided by three specialized agents:

**Agents:** 

1. **Extractor** (explicit limitations from the article), 

2. **Analyzer** (inferred limitations from critical analysis),

3. **Reviewer** (limitations from an open review perspective).

**Goals**:

1. Combine all limitations into a cohesive, non-redundant list. 

2. Ensure each limitation is clearly stated, scientifically valid, and aligned with the article's content.

3. Prioritize critical limitations that affect the paper's validity and reproducibility.

4. Format the final list in a clear, concise, and professional manner, suitable for a scientific review or report.

**Workflow**:

1. **Plan**: Outline how to synthesize limitations, identify potential redundancies, and resolve discrepancies.

2. **Analyze**: Combine limitations, prioritizing critical ones, and verify alignment with the article.

3. **Reflect**: Check for completeness, scientific rigor, and clarity.

4. **Continue**: Iterate until the list is comprehensive, non-redundant, and professionally formatted.

**Output Format**:
- Numbered list of final limitations.

- For each: Clear statement, brief justification, and source in brackets (e.g., [Author-stated], [Inferred], [Peer-review-derived]).

Extractor Agent Analysis:
extractor output

Analyzer Agent Analysis:
analyzer output

Reviewer Agent Analysis:
reviewer output

Please merge these three different perspectives on the paper's limitations into a comprehensive, well-organized analysis. Synthesize the insights, resolve any contradictions, and provide a unified view of the paper's limitations.

              \texttt{\textquotesingle\textquotesingle\textquotesingle}
            \end{minipage}
        };
    \end{tikzpicture}
    \caption{Master Agent}
    \label{fig:master-agent}
\end{figure*}

\begin{figure*}[htbp]
    \centering
    \begin{tikzpicture}
        \node[draw, rectangle, rounded corners, fill=yellow!20, text width=0.85\textwidth, inner sep=5pt] (box) {
            \begin{minipage}{\textwidth}
                \small
                \raggedright 
                \textbf{Prompt} = \texttt{\textquotesingle\textquotesingle\textquotesingle} \\
You are a Judge Agent, an expert in evaluating scientific text quality with a focus on limitation generation for
scientific articles. Your task is to assess the outputs of four agents—Extractor (explicit limitations from the article), Analyzer
(inferred limitations from critical analysis), Reviewer (peer-review limitations), and Citation (limitations based on cited papers).
For each agent’s output, assign a numerical score (0–100) and provide specific feedback based on defined criteria. The evaluation is
reference-free, relying on the output’s inherent quality and alignment with each agent’s role.

Evaluation Criteria:
Depth: How critical and insightful is the limitation? Does it reveal significant issues in the study’s design, findings, or implications?
(20

Originality: Is the limitation a generic critique or a novel, context-specific insight? (20

Actionability: Can researchers realistically address the limitation in future work? Does it provide clear paths for improvement?
(30

Topic Coverage: How broadly does the set of limitations cover relevant aspects (e.g., methodology, scope for Extractor/Analyzer; peer
review standards for Reviewer; cited paper gaps for Citation)? (30

Workflow: Plan: Review each agent’s role and expected output (Extractor: explicit limitations; Analyzer: inferred methodological gaps;
Reviewer: peer-review critiques; Citation: cited paper gaps). Identify tools (e.g., text analysis, citation lookup) to verify content
if needed.

Reasoning: Let’s think step by step to evaluate each output: Step 1: Read the agent’s output and confirm its alignment with the agent’s
role. Step 2: Assess each criterion (Depth, Originality, Actionability, Topic Coverage), noting strengths and weaknesses. Step 3: Assign
a score (0–10) for each criterion based on quality, then calculate the weighted total (0–100). Step 4: Generate feedback for each
criterion, specifying what was done well and what needs improvement. Step 5: Verify the evaluation by cross-checking with the article
or cited papers using tools, if necessary.

Analyze: Use tools to verify article or cited paper content to ensure accurate evaluation (e.g., confirm Extractor’s quotes, Citation’s
references). Reflect: Ensure the score and feedback are fair, consistent, and actionable. Re-evaluate if any criterion seems misjudged.
Continue: Iterate until the evaluation is complete for all agents.

Tool Use: Use text analysis tools to verify article content (e.g., Extractor’s quotes, Analyzer’s methodology). Use citation lookup
tools to confirm cited paper details (e.g., Citation’s references). Use web/X search tools to validate Reviewer’s external context,
if needed.

Chain of Thoughts: Document the evaluation process explicitly. For example: “The Extractor’s output identifies a limitation but
lacks critical insight, reducing Depth.” “The Analyzer’s limitation is generic, affecting Originality.” “The Reviewer’s output is
actionable but misses ethical considerations, limiting Topic Coverage.” This narrative ensures transparency and justifies the score
and feedback.

Scoring: For each criterion, assign a score (0–10) based on quality: 0–3: Poor (major issues, e.g., superficial, generic, not actionable,
narrow coverage). 4–6: Fair (moderate issues, e.g., somewhat insightful, partially actionable, incomplete coverage). 7–8: Good
(minor issues, e.g., mostly critical, slightly generic, broadly actionable). 9–10: Excellent (no issues, e.g., highly insightful,
novel, clearly actionable, comprehensive coverage).

Calculate the total score: Sum (criterion score × weight), where weights are Depth (0.2), Originality (0.2), Actionability (0.3),
Topic Coverage (0.3).

Example: Depth (8 × 0.2 = 1.6), Originality (7 × 0.2 = 1.4), Actionability (9 × 0.3 = 2.7), Topic Coverage (6 × 0.3 = 1.8),
Total = (1.6 + 1.4 + 2.7 + 1.8) × 10 = 75.

Input:
Extractor Agent: [extractor agent output]
Analyzer Agent: [analyzer agent output]
Reviewer Agent: [reviewer agent output]
Citation Agent: [citation agent output]

Output Format: The output must strictly be in JSON format, starting with 
For each agent (Extractor, Analyzer, Reviewer, Citation), provide a JSON object with the following structure:

  "agent": "[Agent Name]",
  "total score": [Numerical score, 0–100],
  
  1."evaluation": {
    "Depth": {
      "score": [0–10],
      "strengths": "[What was done well]",
      "issues": "[Problems identified]",
      "suggestions": "[How to improve]"
    },
    
    2."Originality": {
      "score": [0–10],
      "strengths": "[What was done well]",
      "issues": "[Problems identified]",
      "suggestions": "[How to improve]"
    },
    
    3."Actionability": {
      "score": [0–10],
      "strengths": "[What was done well]",
      "issues": "[Problems identified]",
      "suggestions": "[How to improve]"
    },
    
    4."Topic Coverage": {
      "score": [0–10],
      "strengths": "[What was done well]",
      "issues": "[Problems identified]",
      "suggestions": "[How to improve]"
    }
  }
\texttt{\textquotesingle\textquotesingle\textquotesingle}
\end{minipage}
};
    \end{tikzpicture}
    \caption{Judge Agent}
    \label{fig:judge-agent}
\end{figure*}

\begin{figure*}[htbp]
    \centering
    \begin{tikzpicture}
        \node[draw, rectangle, rounded corners, fill=yellow!20, text width=0.85\textwidth, inner sep=5pt] (box) {
            \begin{minipage}{\textwidth}
                \small
                \raggedright 
                \textbf{Prompt} = \texttt{\textquotesingle\textquotesingle\textquotesingle} \\
You are an expert reviewer. Evaluate the quality of the generated limitations based on the following three criteria: Faithfulness,
Soundness, and Importance. For each criterion, assign a score between 1 and 5 and provide a short justification.

Faithfulness = The generated limitations should accurately represent the paper’s content and findings, avoiding any introduction
of misinformation or contradictions to the original concepts, methodologies or results presented.
– 5 points: Perfect alignment with the original content and findings, with no misinformation or contradictions. Fully reflects the
paper’s concepts, methodologies, and results
accurately.
– 4 points: Mostly aligns with the original content but contains minor inaccuracies or slight
misinterpretations. These do not significantly
affect the overall understanding of the paper’s
concepts or results.
– 3 points: Generally aligns with the original
content but includes several minor inaccuracies or contradictions. Some elements may
not fully reflect the paper’s concepts or results,
though the overall understanding is mostly intact.
– 2 points: Noticeable misalignment with the
original content, with multiple inaccuracies
or contradictions that could mislead readers.
Some key aspects of the paper’s concepts or
results are misrepresented.
– 1 point: Introduces significant misalignment
by misrepresenting issues that do not exist in
the paper. Creates considerable misinformation and contradictions that distort the original
content, concepts, or results.

Soundness = The generated limitations should be detailed and specific, with suggestions or critiques that are practical, logically
coherent, and purposeful. It should clearly address relevant aspects of the paper and offer insights that can genuinely improve the
research.
– 5 points: Highly detailed and specific, with
practical, logically coherent, and purposeful
suggestions. Clearly addresses relevant aspects and offers insights that substantially improve the research.
– 4 points: Detailed and mostly specific, with
generally practical and logically sound suggestions. Addresses relevant aspects well but may
lack depth or novelty in some areas.
– 3 points: Detailed and specific but with some
issues in practicality or logical coherence. Suggestions are somewhat relevant and offer partial improvements.
– 2 points: Somewhat vague or lacking in specificity, with suggestions that have limited practicality or logical coherence. Addresses
relevant aspects only partially and provides minimal improvement.
– 1 point: Lacks detail and specificity, with impractical or incoherent suggestions. Fails to
effectively address relevant aspects or offer
constructive insights for improvement.

Importance =  The generated limitations should
address the most significant issues that impact the
paper’s main findings and contributions. They
should highlight key areas where improvements
or further research are needed, emphasizing their
potential to enhance the research’s relevance and
overall impact.
– 5 points: Addresses critical issues that substantially impact the paper’s findings and contributions. Clearly identifies major areas for
significant improvement or further research,
enhancing the research’s relevance and overall
impact.
– 4 points: Identifies meaningful issues that contribute to refining the paper’s findings and
methodology. While the impact is notable,
it does not reach the level of fundamentally
shaping future research directions.
– 3 points: Highlights important issues that offer some improvement to the current work but
do not significantly impact future research directions. Provides useful insights for refining
the paper but lacks broader implications for
further study.
– 2 points: Points out limitations with limited
relevance to the paper’s overall findings and
contributions. Suggestions offer marginal improvements but fail to address more substantial
gaps in the research.
– 1 point: Focuses on trivial issues, such as minor errors or overly detailed aspects. Does not
address substantive issues affecting the paper’s
findings or contributions, limiting its overall
relevance and impact.

Input:
Input Paper: [Input Paper]
LLM Generated Limitations: [LLM Generated Limitations]

Please evaluate the **Generated Limitations** based on the **Input Paper Content** and return your response strictly in the following JSON format:

Faithfulness: rating: , explanation:,
Soundness:    rating: explanation: ,
Importance:   rating: , explanation:
\texttt{\textquotesingle\textquotesingle\textquotesingle}
\end{minipage}
};
    \end{tikzpicture}
    \caption{Evalaution Prompt}
    \label{fig:evaluation-prompt}
\end{figure*}


\begin{thebibliography}{00}
\bibitem{b1} Azher, Ibrahim Al, et al. "BAGELS: Benchmarking the Automated Generation and Extraction of Limitations from Scholarly Text. Findings of the Association for Computational Linguistics: EMNLP 2025" arXiv preprint arXiv:2505.18207 (2025).

\bibitem{b2} Wei, Jason, et al. "Chain-of-thought prompting elicits reasoning in large language models." Advances in neural information processing systems 35 (2022): 24824-24837.  

\bibitem{b3} James Zou and Nitya Thakkar. 2025. Leveraging LLM feedback to enhance review quality. [Online]. Available: https://blog.iclr.cc/2025/04/15/leveraging-llm-feedback-to-enhance-review-quality/

\bibitem{b4} AAAI, “AAAI launches AI-Powered Peer Review Assessment System,” AAAI News, 2025. [Online]. Available: https://aaai.org/aaai-launches-ai-powered-peer-review-assessment-system/


\bibitem{b5} Montori, Victor M., et al. "Users' guide to detecting misleading claims in clinical research reports." Bmj 329.7474 (2004): 1093-1096.

\bibitem{b6} Du, Jiangshu, et al. "LLMs assist NLP researchers: Critique paper (meta-) reviewing." arXiv preprint arXiv:2406.16253 (2024). 

\bibitem{b7} Ross, Paula T., and Nikki L. Bibler Zaidi. "Limited by our limitations." Perspectives on medical education 8.4 (2019): 261-264. 

\bibitem{b8} Ye, Rui, et al. "Are we there yet? revealing the risks of utilizing large language models in scholarly peer review." arXiv preprint arXiv:2412.01708 (2024).


\bibitem{b10} Khabsa, Madian, Pucktada Treeratpituk, and C. Lee Giles. "Ackseer: a repository and search engine for automatically extracted acknowledgments from digital libraries." Proceedings of the 12th ACM/IEEE-CS joint conference on Digital Libraries. 2012. 

\bibitem{b11} Houngbo, Hospice, and Robert E. Mercer. "Method mention extraction from scientific research papers." Proceedings of COLING 2012. 2012. 

\bibitem{b12} Azher, Ibrahim Al, et al. "Futuregen: Llm-rag approach to generate the future work of scientific article." arXiv preprint arXiv:2503.16561 (2025). 

\bibitem{b13} Azher, Ibrahim Al, et al. "Limtopic: Llm-based topic modeling and text summarization for analyzing scientific articles limitations." Proceedings of the 24th ACM/IEEE Joint Conference on Digital Libraries. 2024. 

\bibitem{b14} Al Azher, Ibrahim, and Hamed Alhoori. "Mitigating visual limitations of research papers." 2024 IEEE International Conference on Big Data (BigData). IEEE, 2024. 


\bibitem{b16} Faizullah, Abdur Rahman Bin Mohammed, Ashok Urlana, and Rahul Mishra. "Limgen: Probing the llms for generating suggestive limitations of research papers." Joint European Conference on Machine Learning and Knowledge Discovery in Databases, 2024. 

\bibitem{b17} Xu, Zhijian, et al. "Can LLMs Identify Critical Limitations within Scientific Research? A Systematic Evaluation on AI Research Papers." arXiv preprint arXiv:2507.02694 (2025). 

\bibitem{b18} Ioannidis, John PA. "Limitations are not properly acknowledged in the scientific literature." Journal of clinical epidemiology 60.4 (2007): 324-329. 

\bibitem{b19} Ter Riet, Gerben, et al. "All that glitters isn't gold: a survey on acknowledgment of limitations in biomedical studies." PloS one 8.11 (2013): e73623.

\bibitem{b20} Liu, Zijun, et al. "Dynamic llm-agent network: An llm-agent collaboration framework with agent team optimization." arXiv preprint arXiv:2310.02170 (2023). 

\bibitem{b21} Cemri, Mert, et al. "Why do multi-agent llm systems fail?." arXiv preprint arXiv:2503.13657 (2025). 

\bibitem{b22} Zhou, Ruiyang, Lu Chen, and Kai Yu. "Is LLM a reliable reviewer? a comprehensive evaluation of LLM on automatic paper reviewing tasks." Proceedings of the 2024 LREC-COLING 2024. 

\bibitem{b23} Ye, Rui, et al. "Are we there yet? revealing the risks of utilizing large language models in scholarly peer review." arXiv preprint arXiv:2412.01708 (2024). 

\bibitem{b24} Jin, Yiqiao, et al. "Agentreview: Exploring peer review dynamics with llm agents." arXiv preprint arXiv:2406.12708 (2024). 

\bibitem{b25} Yang, Hui, Sifu Yue, and Yunzhong He. "Auto-gpt for online decision making: Benchmarks and additional opinions." arXiv preprint arXiv:2306.02224 (2023).


\bibitem{b28} Lin, Chin-Yew. "Rouge: A package for automatic evaluation of summaries." Text summarization branches out. 2004.

\bibitem{b29} Hyland, Ken. "Writing without conviction? Hedging in science research articles." Applied linguistics 17.4 (1996): 433-454.


\bibitem{b31} Idahl, Maximilian, and Zahra Ahmadi. "Openreviewer: A specialized large language model for generating critical scientific paper reviews." arXiv preprint arXiv:2412.11948 (2024).

\bibitem{b32} Du, Jiangshu, et al. "LLMs assist NLP researchers: Critique paper (meta-) reviewing." arXiv preprint arXiv:2406.16253 (2024).

\bibitem{b33} Zhu, Minjun, et al. "Deepreview: Improving llm-based paper review with human-like deep thinking process." arXiv preprint arXiv:2503.08569 (2025).

\bibitem{b34} Wang, Qingyun, et al. "ReviewRobot: Explainable paper review generation based on knowledge synthesis." arXiv preprint arXiv:2010.06119 (2020).

\bibitem{b35} Robertson, Zachary. "Gpt4 is slightly helpful for peer-review assistance: A pilot study." arXiv preprint arXiv:2307.05492 (2023).

\bibitem{b36} Liang, Weixin, et al. "Can large language models provide useful feedback on research papers? A large-scale empirical analysis." NEJM AI 1.8 (2024): AIoa2400196. 

\bibitem{b37} Kuznetsov, Ilia, et al. "Revise and resubmit: An intertextual model of text-based collaboration in peer review." Computational Linguistics 48.4 (2022): 949-986.

\bibitem{b38} Tan, Cheng, et al. "Peer review as a multi-turn and long-context dialogue with role-based interactions." arXiv preprint arXiv:2406.05688 (2024).

\bibitem{b39} Azher, Ibrahim Al. "Generating suggestive limitations from research articles using llm and graph-based approach." Proceedings of the 24th ACM/IEEE Joint Conference on Digital Libraries. 2024.


\bibitem{b40} J. Zou et al., "Assisting ICLR 2025 reviewers with feedback," ICLR Blog, Oct. 9, 2024. [Online]. Available: https://blog.iclr.cc/2024/10/09/iclr2025-assisting-reviewers/ 

\bibitem{b41} D'Arcy, Mike, et al. "Marg: Multi-agent review generation for scientific papers." arXiv preprint arXiv:2401.04259 (2024).


\end{thebibliography}
\end{document}